\documentclass[journal]{vgtc}                     
\usepackage{amssymb}
\usepackage[svgnames]{xcolor} 
\usepackage[normalem]{ulem}
\usepackage{tabularray}
\usepackage{array}
\usepackage{ragged2e}
\usepackage{colortbl}
\usepackage{booktabs}
\usepackage{graphicx}
\usepackage{tabularx}
\usepackage{amssymb}
\usepackage[untagged, flatstructure]{accessibility}

\onlineid{0}

\vgtccategory{Research}

\vgtcpapertype{evaluation}

\title{Can GPT-4 Models Detect Misleading Visualizations?}

\usepackage{tikz}

\author{%
  Jason Alexander,
  Priyal Nanda, Kai-Cheng Yang, and 
  Ali Sarvghad
}

\authorfooter{
  \item
  	Jason Alexander is with University of Massachusetts Amherst.
  	E-mail: jasonhuangalexander@gmail.com.
  \item
  	Priyal Nanda is with University of Massachusetts Amherst.
  	E-mail: phnanda@umass.edu.

  \item Kai-Cheng Yang is with Network Science Institute, Northeastern University.
  	E-mail: yang3kc@gmail.com.

   \item Ali Sarvghad is with University of Massachusetts Amherst.
  	E-mail: asarv@cs.umass.edu.
}

\abstract{%
The proliferation of misleading visualizations online, particularly during critical events like public health crises and elections, poses a significant risk.
This study investigates the capability of GPT-4 models (4V, 4o, and 4o mini) to detect misleading visualizations.
Utilizing a dataset of tweet-visualization pairs containing various visual misleaders, we test these models under four experimental conditions with different levels of guidance.
We show that GPT-4 models can detect misleading visualizations with moderate accuracy without prior training (naive zero-shot) and that performance notably improves when provided with definitions of misleaders (guided zero-shot).
However, a single prompt engineering technique does not yield the best results for all misleader types.
Specifically, providing the models with misleader definitions and examples (guided few-shot) proves more effective for reasoning misleaders, while guided zero-shot performs better for design misleaders.
This study underscores the feasibility of using large vision-language models to detect visual misinformation and the importance of prompt engineering for optimized detection accuracy. 
}

\usepackage{tabu}                      
\usepackage{booktabs}                  
\usepackage{lipsum}                    
\usepackage{mwe}                       
\usepackage{soul}

\newcommand{\plusicon}{~\includegraphics[height=1.5ex]{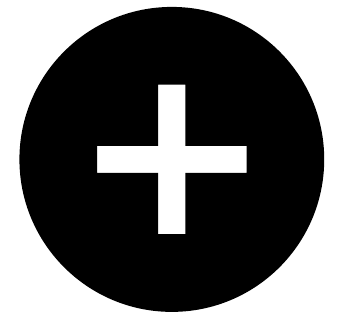}\xspace}
\newcommand{\oneicon}{~\includegraphics[height=1.7ex]{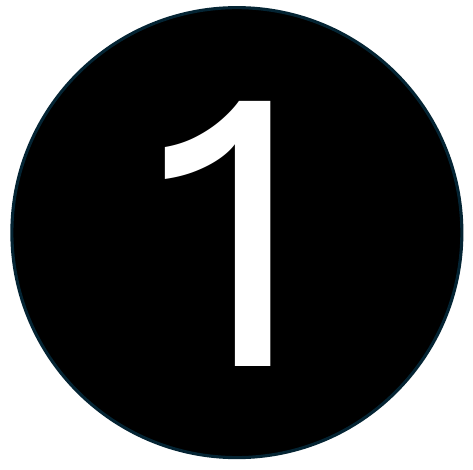}\xspace}
\newcommand{\twoicon}{~\includegraphics[height=1.7ex]{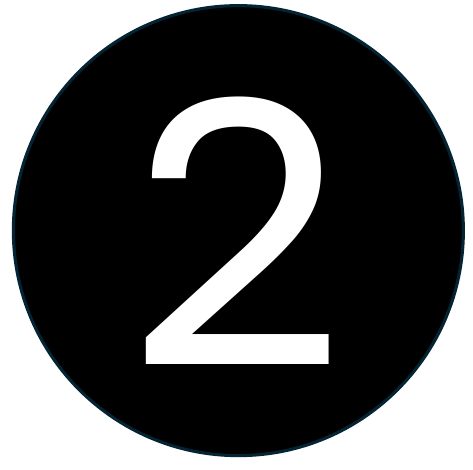}\xspace}
\newcommand{\threeicon}{~\includegraphics[height=1.7ex]{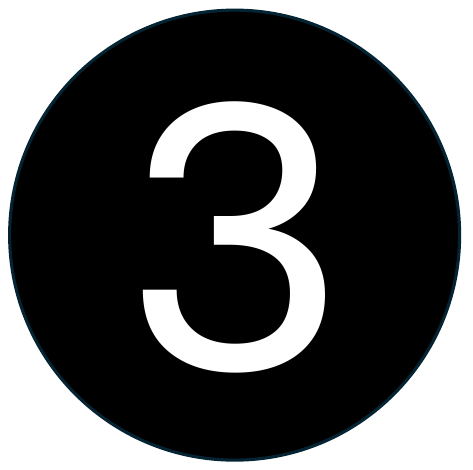}\xspace}

\newenvironment{tight_itemize}{\begin{itemize} \itemsep -1pt}{\end{itemize}}

\teaser{
  \centering
  \includegraphics[width=\linewidth]{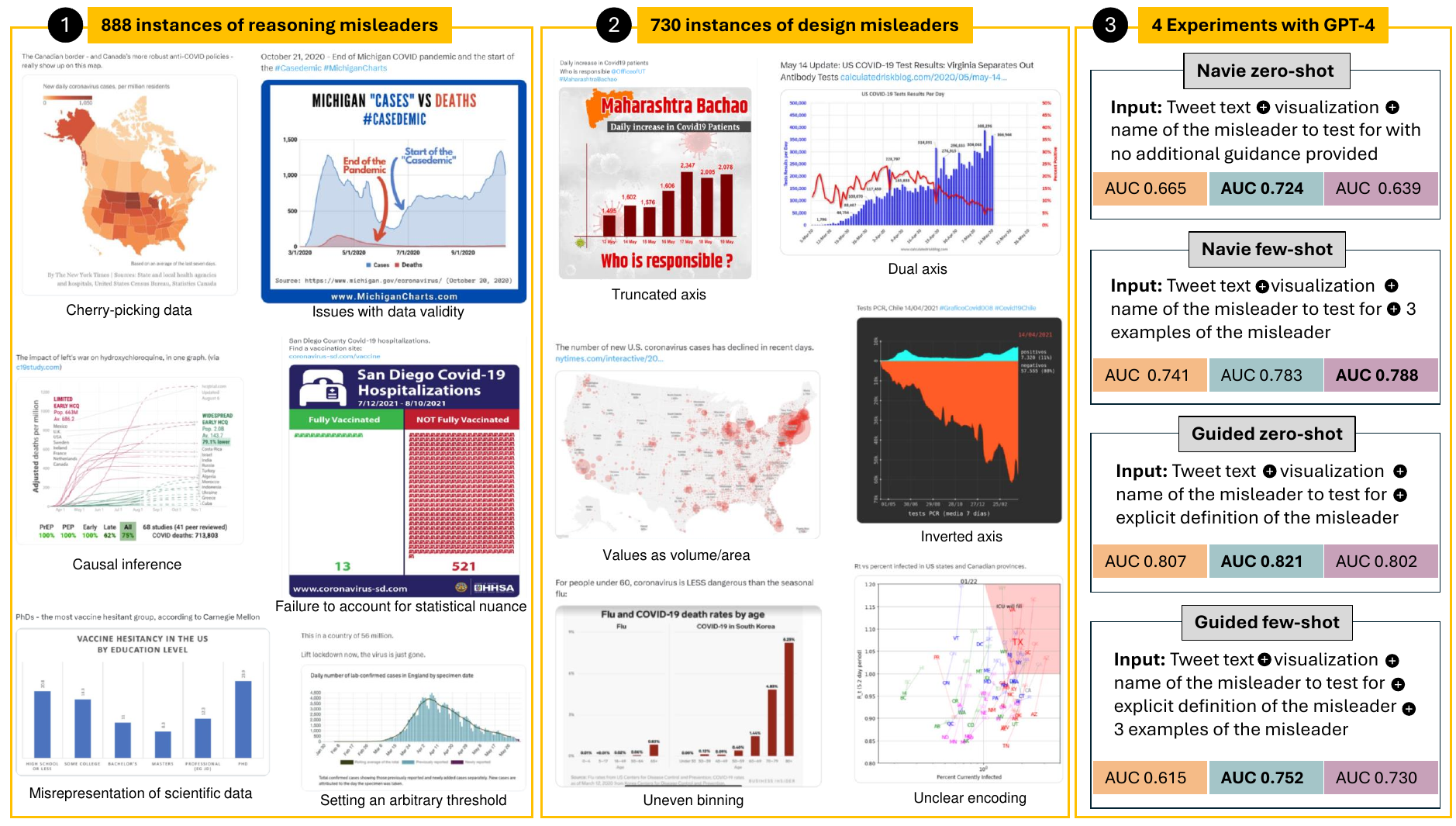}
  \alt{This figure is divided into three main sections: (1) Instances of Reasoning Misleaders, which contains 888 instances, with examples including a map of North America with cherry-picking data, a graph titled ``Michigan `CASES' vs DEATHS \#CASEDEMIC'' highlighting issues with data validity, and various charts demonstrating causal inference, statistical nuanced issues, and arbitrary thresholds. (2) Instances of Design Misleaders, containing 730 instances, with examples such as a bar chart with a dual axis titled ``Who is responsible?'', a graph with an inverted axis showing COVID-19 cases, and a scatter plot with unclear encoding. (3) Experiments with LVLMs, detailing four types of experiments conducted with three models (GPT-4o mini, GPT-4o, GPT-4V): Naive Zero-Shot, Naive Few-Shot, Guided Zero-Shot, and Guided Few-Shot, each with different input configurations and AUC scores. Naive Zero-Shot involves tweet text, visualization, and misleader name without guidance (AUC scores 0.665, 0.724, 0.639); Naive Few-Shot adds three examples (AUC scores 0.741, 0.783, 0.788); Guided Zero-Shot includes explicit definitions (AUC scores 0.807, 0.821, 0.802); and Guided Few-Shot combines all elements (AUC scores 0.615, 0.752, 0.730). The caption highlights that this work examines the accuracy of three OpenAI models in detecting misleading visualizations, with the Guided Zero-Shot setup yielding the best overall performance (AUC=0.821), followed by Naive Few-Shot (AUC=0.788). The performance of GPT-4o and GPT-4V was comparable and slightly better than GPT-4o mini. The dataset used is a subset of previously curated and labeled instances by Lisnic et al.}
  \caption{In this work, we examined the accuracy of three OpenAI models, GPT-4o mini (\includegraphics[height=1.5ex]{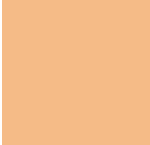}), GTP-4o (\includegraphics[height=1.5ex]{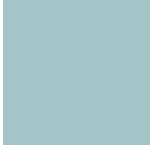}), and GPT-4V (\includegraphics[height=1.5ex]{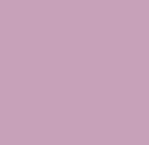}) in detecting misleading visualizations using a test dataset that includes \oneicon 888 instances of reasoning and \twoicon 730 instances of design misleaders (examples in the figure). \threeicon We performed four experiments, gradually increasing the level of guidance provided as input to the models to assist their reasoning and performance. Guided zero-shot yielded the overall best performance across all three models and guidance levels (AUC=0.821 \includegraphics[height=1.5ex]{figs/light-blue.pdf}) followed by naive few-shot (AUC=0.788 \includegraphics[height=1.5ex]{figs/light-purple.pdf}). The performance of GPT-4o and GPT-4V are comparable and slightly better than GPT-4o mini. Our dataset is a subset of that previously curated and labeled for various misleaders by Lisnic et al.~\cite{lisnic2023misleading}.
  }
  \label{fig:teaser}
}

\usepackage{mathptmx}                  

\begin{document}



\firstsection{Introduction}
\label{sec:introduction}
\maketitle

Misleading visualizations are graphical depictions of data that can influence a viewer's perception and judgment to provoke specific inferences and conclusions~\cite{lisnic2023misleading,lo2022misinformed}.
They can be constructed by manipulating data (e.g., cherry-picking), visualization design (e.g., truncated axes), or accompanying content (e.g., biased caption).
The digital era and social media have notably amplified their spread and reach.
Remarkably, false information spreads faster and more broadly than factual information on social media, suggesting that misleading visualizations are also likely to spread widely, reaching considerable audiences~\cite{vosoughi2018spread, lisnic2023yeah, cairo2019charts, shao2018spread}.
Like other forms of misinformation, misleading visualizations can have adverse personal and social consequences.
For example, during the COVID-19 pandemic, exposure to misleading visualizations and misinformation reduced some people's willingness to vaccinate~\cite{loomba2021measuring} and comply with other preventative measures such as wearing masks~\cite{pennycook2020fighting}.

Enabling everyday consumers of online information to detect and avoid misleading visualizations is a non-trivial task. Examining the veracity of information and visualizations requires complex knowledge, including sufficient data visualization, media, and scientific literacy~\cite{lin2021fooled,rivera2022social, lo2022misinformed, langraw2023study, vrabec2023popularisation} as well as critical analysis skills~\cite{brown2023social}, which many everyday users of digital social platforms may lack. 
Hence, education is often regarded as the primary defense against misleading visualizations ~\cite{jones2004models, galesic2010statistical, cairo2019charts, bergstrom2021calling, van2017inoculating}.
This countermeasure alone, however, may not be adequate.
The visual representation of data carries an implicit authority that many viewers accept at face value without further scrutiny~\cite{correll2019ethical,kennedy2016work, cairo2019charts}.  People may also fail to detect misleaders due to psychological and cognitive factors that inhibit attention to details and scrutiny, such as inattentional blindness~\cite{simons1999gorillas}, strained cognitive resources~\cite{sweller1988cognitive}, and the overconfidence effect~\cite{kruger1999unskilled} [in their visualization literacy].

Automatic detection of misleading visualizations can potentially complement interventions such as education.
In particular, large vision-language models (LVLMs) offer new opportunities due to their potential for analyzing and reasoning about visualizations~\cite{kim2023good, vazquez2024llms, do2023llms} and enhancing the detection of harmful online memes and misinformation~\cite{lin2023beneath, vergho2024comparing, brahma2023leveraging,zakir2023infodemics, zhang2023detecting}.
However, the effectiveness and utility of LVLMs remain to be thoroughly examined.

As a first step in this direction, we examined the ability and accuracy of three GPT-4 models, i.e., 4V~\cite{GPT4Visi34:online}, 4o~\cite{GPT4o:online}, and 4o mini~\cite{GPT4omini:online}, a family of LVLMs from OpenAI with the ability to interpret and respond to both text and image inputs in detecting visualization misleaders.
Specifically, we explored two critical research questions: \textbf{(RQ1)} How accurately can GPT-4 models detect misleaders using zero-shot learning? and  \textbf{(RQ2)} Can prompt engineering techniques, such as few-shot learning, enhance its performance? RQ1 examines the innate ability of GPT-4 models in reasoning about and detecting misleaders. RQ2 gauges possible performance gains by providing the models with guidance.

We prepared a test dataset with 1,618 positive (i.e., with misleaders) and 1,618 negative (i.e., without misleaders) samples of tweet-visualization pairs, derived from the dataset created by Lisnic et al. containing 9,958 labeled visualizations~\cite{lisnic2023misleading}. 
Figure~\ref{fig:teaser} provides examples of positive samples tested in our study. Next, we selected seven design and seven reasoning visualization misleaders, detailed in Table~\ref{tab:breakdown-analysis} and Section~\ref{sec:experiment_design}, from Lisnic et al.~\cite{lisnic2023misleading}. Subsequently, we assessed GPT-4 models' performance under four setups, controlling the level of guidance provided to the model from no guidance to rich guidance with misleader definitions and examples:
\begin{tight_itemize}
    \item \textbf{Naive zero-shot:} tweet text\plusicon visualization\plusicon name of the misleader to be tested for with no additional examples or guidance 
    
    \item \textbf{Naive few-shot:} tweet text\plusicon visualization\plusicon name of the misleader to be tested for\plusicon 3 examples of the misleader
     
     \item \textbf{Guided zero-shot:} tweet text\plusicon visualization\plusicon name of the misleader \plusicon explicit definition of the misleader and instruction
    
    \item \textbf{Guided few-shot:} tweet text\plusicon visualization\plusicon name of the misleader\plusicon 3 examples of the misleader\plusicon explicit definition of the misleader and instruction
\end{tight_itemize}

We evaluated the performance of tested models in detecting misleaders using the Area Under the Receiver Operating Characteristic Curve (ROC-AUC), a metric that quantifies the ability of a binary classification model to distinguish between classes. The AUC values range from 0 to 1, with 1 indicating perfect classification and 0.5 indicating random guessing. The AUC scores (\autoref{fig:main_results}) for the three tested models were consistently above the 0.5 threshold across all conditions, demonstrating the feasibility of using GPT-4 models and perhaps comparable LVLMs for detecting misleading visualizations (\textbf{RQ1}). Among the tested models, GPT-4o offered the best overall performance. Our results (\autoref{tab:breakdown-analysis}) also demonstrated that providing guidance improved the tested models' performance (\textbf{RQ2}), though the gain varied across misleaders. For some misleaders, such as ``Causal inference,'' ``Cherry-picking,'' and ``Dual axis,'' the model's performance was consistently high regardless of the guidance levels. In contrast, the accuracy in detecting some misleaders notably increased with more guidance. For example, the performance in detecting ``Setting an arbitrary threshold'' and ``Incorrect reading of chart'' improved notably from naive zero-shot (AUC=0.241 and 0.540) to guided few-shot (AUC=0.806 and 0.660). Our findings reveal the relationship between the type of misleader and the guidance setup that improves detection. Identifying reasoning misleaders appeared to benefit more from detailed examples and explicit definitions provided by the guided few-shot setup, whereas the best overall accuracy in detecting design misleaders was achieved under the guided zero-shot setup. This suggests that less guidance is beneficial for identifying design misleaders compared to reasoning misleaders.

\section{Related Work}
\label{sec:related_work}

\subsection{Misleading Visualizations}
Misleading visualizations, although not new~\cite{lo2022misinformed}, have been extensively categorized and analyzed in recent research. Lo et al.~\cite{lo2022misinformed} identified 74 types of misleading issues across over 1,000 visualizations, establishing a comprehensive taxonomy. Lisnic et al.~\cite{lisnic2023misleading} examined 9,958 tweet-visualization pairs related to COVID-19, discussing manipulative design and reasoning techniques. These studies highlight the ongoing need to investigate how visualizations can deceive, the underlying mechanisms, and identification and mitigation strategies.

\subsection{ML for Combating Misleading Visualizations}
The use of machine learning (ML) in detecting and addressing misleading visualizations has evolved significantly, with various approaches being applied to enhance the reliability and accuracy of visual content analysis. These ML techniques range from basic image recognition to complex data integration analysis within visual content.

One prominent approach has been the use of Convolutional Neural Networks (CNNs), renowned for their effectiveness in image classification and analysis.
CNNs have been extensively applied in projects like ChartSense\cite{jung2017chartsense}, where ML models assist in the interactive classification of and information extraction from images, which is crucial for verifying the accuracy of the information portrayed in these charts. The integration of natural language processing (NLP) with visual analysis to provide a more holistic content assessment has also gained traction.
For instance, the CHARTVE model~\cite{huang2023lvlms} uses a blend of visual cues and text analysis to detect inconsistencies or factual errors in chart captions.

While effective, these methods require substantial resources and often struggle with generalizability~\cite{do2023llms}.
For instance, to ensure the effectiveness of supervised ML models, they must be continuously trained on freshly labeled (often by humans) data.

Our project extends the use of GPT-4 models beyond text analysis, employing advanced prompt engineering and methodologies like the TReE framework~\cite{yang2024empowering}. These techniques enhance GPT-4 models' ability to identify and evaluate visual misinformation. Our research aims to set a new standard for applying pre-trained models in digital media literacy, ensuring more reliable media consumption.

\section{Test Dataset and Experiments}
\label{sec:experiment_design}

We derived our dataset from Lisnic et al.~\cite{lisnic2023misleading} dataset, consisting of 9,958 tweet-visualization pairs.
Misleaders (or their lack thereof) in these visualizations were manually labeled.
The authors grouped misleaders into two categories: \textit{Design Misleaders} and \textit{Reasoning Misleaders}.
Visualizations in the first category contain visual design and construction that primarily trigger misinterpretation and misinformation (e.g., truncated Y axis).
Whereas instances in the second category feature manipulation of data and accompanying visualization content, leading to biases and misjudgments (e.g., cherry-picking).
Figure~\ref{fig:teaser} shows instances of design and reasoning misleaders. 

\subsection{Preparation of Test Dataset}
We began preparing our test set with a spreadsheet shared by Lisnic et al.~\cite{lisnic2023misleading} containing IDs and associated misleader labels for 9,958 tweets.
We successfully collected 9,042 tweets from Twitter (now X), while the rest were unavailable due to removal, privacy settings, or technical issues.
We excluded 1,992 tweets involved in conversations and 216 in threads to reduce the impact of extraneous information on our experiments.
After downloading images for the remaining tweets, we removed 41 tweets with zero-byte images and 809 tweets with multiple images, resulting in 5,984 candidate tweets for testing.

\begin{table*}
\centering
\setlength{\extrarowheight}{0pt}
\addtolength{\extrarowheight}{\aboverulesep}
\addtolength{\extrarowheight}{\belowrulesep}
\setlength{\aboverulesep}{0pt}
\setlength{\belowrulesep}{0pt}
\alt{This table presents the AUC scores for different types of misleaders, divided into reasoning and design misleaders, and evaluated across four experimental setups (Naive Zero-Shot, Naive Few-Shot, Guided Zero-Shot, Guided Few-Shot) using three models (GPT-4o mini, GPT-4o, GPT-4V). The table includes the number of positive and negative samples for each misleader type.

In the reasoning misleaders section, the misleaders include Causal Inference, Cherry-Picking, Setting an Arbitrary Threshold, Issues with Data Validity, Failure to Account for Statistical Nuance, Misrepresentation of Scientific Studies, and Incorrect Reading of Chart. For example, in the Naive Zero-Shot setup, the AUC score for Causal Inference is 0.816 for GPT-4o mini, 0.806 for GPT-4o, and 0.534 for GPT-4V. The Guided Few-Shot setup shows improved performance with an AUC score of 0.872 for GPT-4o in the Causal Inference category.

In the design misleaders section, the misleaders include Value as Area/Volume, Dual Axis, Truncated Axis, Inverted Axis, Unclear Encoding, Inappropriate Encoding, and Uneven Binning. For instance, in the Naive Zero-Shot setup, the AUC score for Dual Axis is 0.903 for GPT-4o mini, 0.906 for GPT-4o, and 0.836 for GPT-4V. The Guided Zero-Shot setup shows significant improvement with an AUC score of 0.970 for GPT-4o in the Dual Axis category.

The table also includes subtotal and total AUC scores for reasoning and design misleaders, demonstrating overall trends in model performance across different setups. For example, the total AUC score for the Guided Zero-Shot setup is 0.807, indicating the highest performance across all models and misleader types compared to other setups.}
\caption{Summary statistics of our test dataset and the experiment results with GPT-4o mini (\includegraphics[height=1.5ex]{figs/light-orange.pdf}), GTP-4o (\includegraphics[height=1.5ex]{figs/light-blue.pdf}), and GPT-4V (\includegraphics[height=1.5ex]{figs/light-purple.pdf}). From left to right, we report the misleader names, numbers of positive and negative samples, and AUC scores for four experiment setups. In addition to the results for each misleader, we also report the aggregate AUC scores.  The highest AUC scores for each row are highlighted in \textbf{bold}. Scores equal to or less than 0.5 are marked with \underline{underlines}.}
\label{tab:breakdown-analysis}
\resizebox{\linewidth}{!}{%
\begin{tabular}{|>{\hspace{0pt}}m{0.187\linewidth}|>{\centering\hspace{0pt}}m{0.069\linewidth}|>{\centering\hspace{0pt}}m{0.073\linewidth}|>{\centering\hspace{0pt}}m{0.05\linewidth}>{\centering\hspace{0pt}}m{0.05\linewidth}>{\centering\hspace{0pt}}m{0.05\linewidth}|>{\centering\hspace{0pt}}m{0.05\linewidth}>{\centering\hspace{0pt}}m{0.05\linewidth}>{\centering\hspace{0pt}}m{0.05\linewidth}|>{\centering\hspace{0pt}}m{0.05\linewidth}>{\centering\hspace{0pt}}m{0.05\linewidth}>{\centering\hspace{0pt}}m{0.05\linewidth}|>{\centering\hspace{0pt}}m{0.05\linewidth}>{\centering\hspace{0pt}}m{0.05\linewidth}>{\centering\arraybackslash\hspace{0pt}}m{0.05\linewidth}|} 
\toprule
\textbf{Reasoning misleaders} & \textbf{\# Positive} & \hspace{-4pt}\textbf{\# Negative} & \multicolumn{3}{>{\centering\hspace{-30pt}}m{0.15\linewidth}|}{\textbf{Naive zero-shot}} & \multicolumn{3}{>{\centering\hspace{-30pt}}m{0.15\linewidth}|}{\textbf{\textbf{Naive few-shot}}} & \multicolumn{3}{>{\centering\hspace{-30pt}}m{0.15\linewidth}|}{\textbf{Guided zero-shot}} & \multicolumn{3}{>{\centering\arraybackslash\hspace{-30pt}}m{0.15\linewidth}|}{\textbf{Guided few-shot}} \\ 
\hline
Causal inference & 299 & 299 & {\cellcolor[rgb]{0.961,0.733,0.529}}0.816 & {\cellcolor[rgb]{0.643,0.773,0.784}}0.806 & {\cellcolor[rgb]{0.78,0.631,0.725}}0.534 & {\cellcolor[rgb]{0.961,0.733,0.529}}\textbf{\textbf{0.892}} & {\cellcolor[rgb]{0.643,0.773,0.784}}0.795 & {\cellcolor[rgb]{0.78,0.631,0.725}}0.863 & {\cellcolor[rgb]{0.961,0.733,0.529}}0.848 & {\cellcolor[rgb]{0.643,0.773,0.784}}0.845 & {\cellcolor[rgb]{0.78,0.631,0.725}}0.840 & {\cellcolor[rgb]{0.961,0.733,0.529}}0.844 & {\cellcolor[rgb]{0.643,0.773,0.784}}0.872 & {\cellcolor[rgb]{0.78,0.631,0.725}}0.865 \\
Cherry-picking & 251 & 251 & {\cellcolor[rgb]{0.961,0.733,0.529}}0.748 & {\cellcolor[rgb]{0.643,0.773,0.784}}0.830 & {\cellcolor[rgb]{0.78,0.631,0.725}}0.697 & {\cellcolor[rgb]{0.961,0.733,0.529}}0.804 & {\cellcolor[rgb]{0.643,0.773,0.784}}0.822 & {\cellcolor[rgb]{0.78,0.631,0.725}}0.814 & {\cellcolor[rgb]{0.961,0.733,0.529}}0.785 & {\cellcolor[rgb]{0.643,0.773,0.784}}0.752 & {\cellcolor[rgb]{0.78,0.631,0.725}}\textbf{\textbf{0.839}} & {\cellcolor[rgb]{0.961,0.733,0.529}}0.718 & {\cellcolor[rgb]{0.643,0.773,0.784}}0.813 & {\cellcolor[rgb]{0.78,0.631,0.725}}0.753 \\
Setting an arbitrary threshold & 199 & 199 & {\cellcolor[rgb]{0.961,0.733,0.529}}\uline{0.241} & {\cellcolor[rgb]{0.643,0.773,0.784}}0.351 & {\cellcolor[rgb]{0.78,0.631,0.725}}0.347 & {\cellcolor[rgb]{0.961,0.733,0.529}}0.747 & {\cellcolor[rgb]{0.643,0.773,0.784}}0.671 & {\cellcolor[rgb]{0.78,0.631,0.725}}0.575 & {\cellcolor[rgb]{0.961,0.733,0.529}}0.628 & {\cellcolor[rgb]{0.643,0.773,0.784}}0.614 & {\cellcolor[rgb]{0.78,0.631,0.725}}0.620 & {\cellcolor[rgb]{0.961,0.733,0.529}}0.753 & {\cellcolor[rgb]{0.643,0.773,0.784}}\textbf{\textbf{0.806}} & {\cellcolor[rgb]{0.78,0.631,0.725}}0.618 \\
Issues with data validity & 61 & 61 & {\cellcolor[rgb]{0.961,0.733,0.529}}0.797 & {\cellcolor[rgb]{0.643,0.773,0.784}}0.787 & {\cellcolor[rgb]{0.78,0.631,0.725}}0.606 & {\cellcolor[rgb]{0.961,0.733,0.529}}0.858 & {\cellcolor[rgb]{0.643,0.773,0.784}}0.810 & {\cellcolor[rgb]{0.78,0.631,0.725}}0.672 & {\cellcolor[rgb]{0.961,0.733,0.529}}0.793 & {\cellcolor[rgb]{0.643,0.773,0.784}}0.730 & {\cellcolor[rgb]{0.78,0.631,0.725}}0.698 & {\cellcolor[rgb]{0.961,0.733,0.529}}0.807 & {\cellcolor[rgb]{0.643,0.773,0.784}}\textbf{\textbf{0.877}} & {\cellcolor[rgb]{0.78,0.631,0.725}}0.609 \\
Failure to account for statistical nuance & 59 & 59 & {\cellcolor[rgb]{0.961,0.733,0.529}}0.774 & {\cellcolor[rgb]{0.643,0.773,0.784}}0.763 & {\cellcolor[rgb]{0.78,0.631,0.725}}0.679 & {\cellcolor[rgb]{0.961,0.733,0.529}}0.872 & {\cellcolor[rgb]{0.643,0.773,0.784}}0.796 & {\cellcolor[rgb]{0.78,0.631,0.725}}\textbf{\textbf{\textbf{\textbf{0.896}}}} & {\cellcolor[rgb]{0.961,0.733,0.529}}0.895 & {\cellcolor[rgb]{0.643,0.773,0.784}}0.645 & {\cellcolor[rgb]{0.78,0.631,0.725}}0.762 & {\cellcolor[rgb]{0.961,0.733,0.529}}0.864 & {\cellcolor[rgb]{0.643,0.773,0.784}}0.823 & {\cellcolor[rgb]{0.78,0.631,0.725}}0.824 \\
Misrepresentation of scientific studies & 14 & 14 & {\cellcolor[rgb]{0.961,0.733,0.529}}0.832 & {\cellcolor[rgb]{0.643,0.773,0.784}}0.709 & {\cellcolor[rgb]{0.78,0.631,0.725}}0.571 & {\cellcolor[rgb]{0.961,0.733,0.529}}0.921 & {\cellcolor[rgb]{0.643,0.773,0.784}}0.783 & {\cellcolor[rgb]{0.78,0.631,0.725}}0.676 & {\cellcolor[rgb]{0.961,0.733,0.529}}0.913 & {\cellcolor[rgb]{0.643,0.773,0.784}}0.763 & {\cellcolor[rgb]{0.78,0.631,0.725}}0.778 & {\cellcolor[rgb]{0.961,0.733,0.529}}\textbf{\textbf{0.954}} & {\cellcolor[rgb]{0.643,0.773,0.784}}0.936 & {\cellcolor[rgb]{0.78,0.631,0.725}}0.781 \\
Incorrect reading of chart & 5 & 5 & {\cellcolor[rgb]{0.961,0.733,0.529}}0.540 & {\cellcolor[rgb]{0.643,0.773,0.784}}0.600 & {\cellcolor[rgb]{0.78,0.631,0.725}}0.600 & {\cellcolor[rgb]{0.961,0.733,0.529}}0.540 & {\cellcolor[rgb]{0.643,0.773,0.784}}0.640 & {\cellcolor[rgb]{0.78,0.631,0.725}}\textbf{\textbf{\textbf{\textbf{0.800}}}} & {\cellcolor[rgb]{0.961,0.733,0.529}}0.600 & {\cellcolor[rgb]{0.643,0.773,0.784}}0.540 & {\cellcolor[rgb]{0.78,0.631,0.725}}\uline{0.300} & {\cellcolor[rgb]{0.961,0.733,0.529}}\uline{0.500} & {\cellcolor[rgb]{0.643,0.773,0.784}}0.660 & {\cellcolor[rgb]{0.78,0.631,0.725}}0.600 \\ 
\hline
\textbf{Subtotal} & 888 & 888 & {\cellcolor[rgb]{0.961,0.733,0.529}}0.586 & {\cellcolor[rgb]{0.643,0.773,0.784}}0.636 & {\cellcolor[rgb]{0.78,0.631,0.725}}0.526 & {\cellcolor[rgb]{0.961,0.733,0.529}}0.811 & {\cellcolor[rgb]{0.643,0.773,0.784}}0.763 & {\cellcolor[rgb]{0.78,0.631,0.725}}0.763 & {\cellcolor[rgb]{0.961,0.733,0.529}}0.779 & {\cellcolor[rgb]{0.643,0.773,0.784}}0.733 & {\cellcolor[rgb]{0.78,0.631,0.725}}0.749 & {\cellcolor[rgb]{0.961,0.733,0.529}}0.781 & {\cellcolor[rgb]{0.643,0.773,0.784}}\textbf{\textbf{0.835}} & {\cellcolor[rgb]{0.78,0.631,0.725}}0.745 \\ 
\hline
\multicolumn{1}{|>{\hspace{0pt}}m{0.187\linewidth}}{\textbf{Design misleaders}} & \multicolumn{1}{>{\centering\hspace{0pt}}m{0.069\linewidth}}{} & \multicolumn{1}{>{\centering\hspace{0pt}}m{0.073\linewidth}}{} &  &  & \multicolumn{1}{>{\centering\hspace{0pt}}m{0.05\linewidth}}{} &  &  & \multicolumn{1}{>{\centering\hspace{0pt}}m{0.05\linewidth}}{} &  &  & \multicolumn{1}{>{\centering\hspace{0pt}}m{0.05\linewidth}}{} &  &  &  \\ 
\hline
Value as area/volume & 352 & 352 & {\cellcolor[rgb]{0.961,0.733,0.529}}0.609 & {\cellcolor[rgb]{0.643,0.773,0.784}}0.745 & {\cellcolor[rgb]{0.78,0.631,0.725}}0.756 & {\cellcolor[rgb]{0.961,0.733,0.529}}0.623 & {\cellcolor[rgb]{0.643,0.773,0.784}}0.753 & {\cellcolor[rgb]{0.78,0.631,0.725}}0.855 & {\cellcolor[rgb]{0.961,0.733,0.529}}0.842 & {\cellcolor[rgb]{0.643,0.773,0.784}}\textbf{0.873} & {\cellcolor[rgb]{0.78,0.631,0.725}}0.854 & {\cellcolor[rgb]{0.961,0.733,0.529}}0.497 & {\cellcolor[rgb]{0.643,0.773,0.784}}0.719 & {\cellcolor[rgb]{0.78,0.631,0.725}}0.717 \\
Dual axis & 250 & 250 & {\cellcolor[rgb]{0.961,0.733,0.529}}0.903 & {\cellcolor[rgb]{0.643,0.773,0.784}}0.906 & {\cellcolor[rgb]{0.78,0.631,0.725}}0.836 & {\cellcolor[rgb]{0.961,0.733,0.529}}0.849 & {\cellcolor[rgb]{0.643,0.773,0.784}}0.942 & {\cellcolor[rgb]{0.78,0.631,0.725}}0.882 & {\cellcolor[rgb]{0.961,0.733,0.529}}0.909 & {\cellcolor[rgb]{0.643,0.773,0.784}}\textbf{0.970} & {\cellcolor[rgb]{0.78,0.631,0.725}}0.924 & {\cellcolor[rgb]{0.961,0.733,0.529}}0.598 & {\cellcolor[rgb]{0.643,0.773,0.784}}0.747 & {\cellcolor[rgb]{0.78,0.631,0.725}}0.818 \\
Truncated axis & 62 & 62 & {\cellcolor[rgb]{0.961,0.733,0.529}}0.531 & {\cellcolor[rgb]{0.643,0.773,0.784}}0.623 & {\cellcolor[rgb]{0.78,0.631,0.725}}0.597 & {\cellcolor[rgb]{0.961,0.733,0.529}}\uline{0.463} & {\cellcolor[rgb]{0.643,0.773,0.784}}\textbf{\textbf{0.727}} & {\cellcolor[rgb]{0.78,0.631,0.725}}0.653 & {\cellcolor[rgb]{0.961,0.733,0.529}}0.502 & {\cellcolor[rgb]{0.643,0.773,0.784}}0.645 & {\cellcolor[rgb]{0.78,0.631,0.725}}0.653 & {\cellcolor[rgb]{0.961,0.733,0.529}}0.500 & {\cellcolor[rgb]{0.643,0.773,0.784}}0.581 & {\cellcolor[rgb]{0.78,0.631,0.725}}0.540 \\
Inverted axis & 29 & 29 & {\cellcolor[rgb]{0.961,0.733,0.529}}0.617 & {\cellcolor[rgb]{0.643,0.773,0.784}}0.638 & {\cellcolor[rgb]{0.78,0.631,0.725}}0.612 & {\cellcolor[rgb]{0.961,0.733,0.529}}0.586 & {\cellcolor[rgb]{0.643,0.773,0.784}}0.655 & {\cellcolor[rgb]{0.78,0.631,0.725}}0.603 & {\cellcolor[rgb]{0.961,0.733,0.529}}0.636 & {\cellcolor[rgb]{0.643,0.773,0.784}}\textbf{0.672} & {\cellcolor[rgb]{0.78,0.631,0.725}}0.638 & {\cellcolor[rgb]{0.961,0.733,0.529}}0.630 & {\cellcolor[rgb]{0.643,0.773,0.784}}0.569 & {\cellcolor[rgb]{0.78,0.631,0.725}}0.603 \\
Unclear encoding & 21 & 21 & {\cellcolor[rgb]{0.961,0.733,0.529}}0.764 & {\cellcolor[rgb]{0.643,0.773,0.784}}0.688 & {\cellcolor[rgb]{0.78,0.631,0.725}}0.451 & {\cellcolor[rgb]{0.961,0.733,0.529}}0.642 & {\cellcolor[rgb]{0.643,0.773,0.784}}0.688 & {\cellcolor[rgb]{0.78,0.631,0.725}}0.705 & {\cellcolor[rgb]{0.961,0.733,0.529}}\textbf{\textbf{0.787}} & {\cellcolor[rgb]{0.643,0.773,0.784}}0.739 & {\cellcolor[rgb]{0.78,0.631,0.725}}0.701 & {\cellcolor[rgb]{0.961,0.733,0.529}}0.619 & {\cellcolor[rgb]{0.643,0.773,0.784}}0.619 & {\cellcolor[rgb]{0.78,0.631,0.725}}\uline{0.383} \\
Inappropriate encoding & 12 & 12 & {\cellcolor[rgb]{0.961,0.733,0.529}}0.542 & {\cellcolor[rgb]{0.643,0.773,0.784}}0.604 & {\cellcolor[rgb]{0.78,0.631,0.725}}0.538 & {\cellcolor[rgb]{0.961,0.733,0.529}}\textbf{\textbf{0.663}} & {\cellcolor[rgb]{0.643,0.773,0.784}}0.476 & {\cellcolor[rgb]{0.78,0.631,0.725}}\uline{0.451} & {\cellcolor[rgb]{0.961,0.733,0.529}}0.608 & {\cellcolor[rgb]{0.643,0.773,0.784}}0.538 & {\cellcolor[rgb]{0.78,0.631,0.725}}0.625 & {\cellcolor[rgb]{0.961,0.733,0.529}}0.542 & {\cellcolor[rgb]{0.643,0.773,0.784}}0.625 & {\cellcolor[rgb]{0.78,0.631,0.725}}0.500 \\
Uneven binning & 4 & 4 & {\cellcolor[rgb]{0.961,0.733,0.529}}\textbf{0.750} & {\cellcolor[rgb]{0.643,0.773,0.784}}0.438 & {\cellcolor[rgb]{0.78,0.631,0.725}}0.438 & {\cellcolor[rgb]{0.961,0.733,0.529}}\textbf{\textbf{0.750}} & {\cellcolor[rgb]{0.643,0.773,0.784}}0.500 & {\cellcolor[rgb]{0.78,0.631,0.725}}0.625 & {\cellcolor[rgb]{0.961,0.733,0.529}}0.625 & {\cellcolor[rgb]{0.643,0.773,0.784}}\textbf{0.750} & {\cellcolor[rgb]{0.78,0.631,0.725}}0.562 & {\cellcolor[rgb]{0.961,0.733,0.529}}\uline{0.375} & {\cellcolor[rgb]{0.643,0.773,0.784}}\uline{0.375} & {\cellcolor[rgb]{0.78,0.631,0.725}}0.625 \\ 
\hline
\textbf{Subtotal} & 730 & 730 & {\cellcolor[rgb]{0.961,0.733,0.529}}0.737 & {\cellcolor[rgb]{0.643,0.773,0.784}}0.781 & {\cellcolor[rgb]{0.78,0.631,0.725}}0.761 & {\cellcolor[rgb]{0.961,0.733,0.529}}0.704 & {\cellcolor[rgb]{0.643,0.773,0.784}}0.806 & {\cellcolor[rgb]{0.78,0.631,0.725}}0.828 & {\cellcolor[rgb]{0.961,0.733,0.529}}0.826 & {\cellcolor[rgb]{0.643,0.773,0.784}}\textbf{0.875} & {\cellcolor[rgb]{0.78,0.631,0.725}}0.847 & {\cellcolor[rgb]{0.961,0.733,0.529}}0.539 & {\cellcolor[rgb]{0.643,0.773,0.784}}0.702 & {\cellcolor[rgb]{0.78,0.631,0.725}}0.721 \\ 
\hline
\textbf{Total} & 1,618 & 1,618 & {\cellcolor[rgb]{0.961,0.733,0.529}}0.665 & {\cellcolor[rgb]{0.643,0.773,0.784}}0.724 & {\cellcolor[rgb]{0.78,0.631,0.725}}0.639 & {\cellcolor[rgb]{0.961,0.733,0.529}}0.741 & {\cellcolor[rgb]{0.643,0.773,0.784}}0.783 & {\cellcolor[rgb]{0.78,0.631,0.725}}0.788 & {\cellcolor[rgb]{0.961,0.733,0.529}}0.807 & {\cellcolor[rgb]{0.643,0.773,0.784}}\textbf{0.821} & {\cellcolor[rgb]{0.78,0.631,0.725}}0.802 & {\cellcolor[rgb]{0.961,0.733,0.529}}0.615 & {\cellcolor[rgb]{0.643,0.773,0.784}}0.752 & {\cellcolor[rgb]{0.78,0.631,0.725}}0.730 \\
\bottomrule
\end{tabular}
}
\end{table*}

Among these candidates, we identified 1,618 instances of misleaders across 1,228 unique tweets.
There were more misleader instances than tweets because a single tweet might include multiple design and/or reasoning misleaders.
We treated each misleader instance as a distinct positive sample, yielding 730 and 888 design and reasoning misleader instances, respectively.
We further randomly selected 1,618 unique tweets without any misleaders as negative samples.
The final test set has an equal number of positive and negative samples, sourced from a total of 2,846 unique tweets

Table~\ref{tab:breakdown-analysis} offers additional details about the misleaders and their prevalence in the final test dataset. The discrepancy in the number of positive cases among different types of misleaders mirrors the distribution in the original dataset~\cite{lisnic2023misleading}, suggesting that certain issues may be more common in real-world scenarios. Detailed descriptions of each misleader are provided in the supplementary materials.

\subsection{Experiments Design}\label{sec:exprimental_settings}

We evaluated the models' ability to detect misleaders through four distinct experiments, i.e., naive zero-shot, naive few-shot, guided zero-shot, and guided few-shot, each varying in the level of guidance provided to the model.
The naive zero-shot setup tested models' intrinsic ability to identify misleaders without any prior guidance or training, establishing a baseline for comparing the performance of other setups. The naive few-shot experiment leveraged the models' in-context learning capabilities, which have been shown to enhance performance across various tasks~\cite{brown2020language}.
The guided zero-shot setup included explicit, detailed definitions of misleaders, directing the models to integrate this information into their analysis. We adopted these definitions from Lisnic et al.~\cite{lisnic2023misleading} to align the guidance provided to the model with that used by human annotators who labeled the data initially.
The guided few-shot setup provided the most extensive guidance, combining three examples and the definition of the specific misleader being tested. OpenAI's official prompt engineering guide~\cite{Prompten19:online} recommends few-shot and guided techniques as effective methods to improve model performance. Examples of complete prompts for each experiment and corresponding responses are included in the supplementary materials.

We performed all the experiments by querying OpenAI's API endpoints for the three models (\textit{gpt-4o-mini-2024-07-18}, \textit{gpt-4o-2024-05-13}, and \textit{gpt-4-1106-vision-preview}) while setting the temperature to zero for consistent and deterministic output as well as reproducibility.
The same system prompt, ``You are an expert on analyzing misleading scientific visualizations. Your job is to identify the misleading aspects in the given chart,'' was used across all setups.
For every sample, a model was asked to provide a degree of certainty, from 0 to 100, that the sample tweet suffered from the misleader being tested for and an explanation. We kept the core structure of the prompts consistent across all experiments, only updating the misleader names or adding information as required by the experimental setup.

The prompts used to test for design and reasoning misleaders were distinct. For design misleaders, which rely solely on the construction of the visualization, only images were provided to the models. In contrast, both tweet texts and corresponding images were provided for reasoning misleaders, as the interpretation of the visualizations mattered as well. We issued a total of 38,832 queries to the three models.

\section{Data Analysis and Results}
\label{sec:results}

\begin{figure}
    \centering
    \includegraphics[width=\columnwidth]{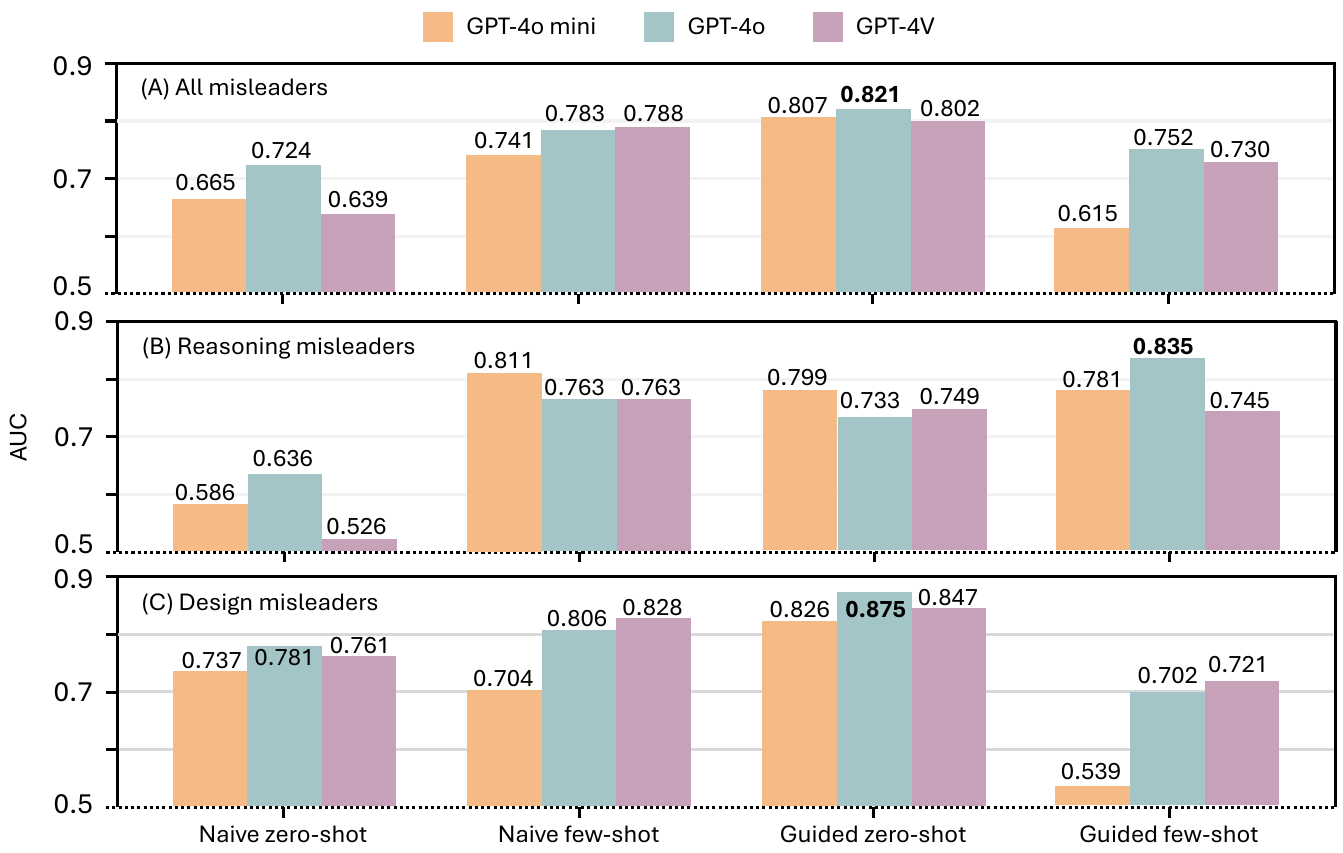}
    \alt{in detecting misleading visualizations across different guidance setups. The figure is divided into three panels: (A) All Misleaders, (B) Reasoning Misleaders, and (C) Design Misleaders. Each panel shows the Area Under the Curve (AUC) scores for Naive Zero-Shot, Naive Few-Shot, Guided Zero-Shot, and Guided Few-Shot experiments. In panel (A), for all misleaders, GPT-4o performed best in the Guided Zero-Shot setup with an AUC of 0.821, followed by Naive Few-Shot with an AUC of 0.788. GPT-4V and GPT-4o mini showed comparable performance with slightly lower AUC scores. In panel (B), for reasoning misleaders, GPT-4o also performed best in the Guided Few-Shot setup with an AUC of 0.835. In panel (C), for design misleaders, GPT-4o achieved the highest performance in the Guided Zero-Shot setup with an AUC of 0.875, while GPT-4V and GPT-4o mini had lower scores. The figure demonstrates that guided setups generally yield better performance, with Guided Zero-Shot being particularly effective for design misleaders and Guided Few-Shot excelling in reasoning misleaders.}
    \caption{
    The overall performance of models across different guidance levels and (A) all, (B) reasoning, and (C) design misleaders. The highest AUC scores are highlighted in \textbf{bold}. The AUC of 0.5 is the random guessing threshold. 
    }
    \label{fig:main_results}
\end{figure}

We used the Area Under the Receiver Operating Characteristic Curve (ROC-AUC) to assess model performance in four different experiment settings.
AUC scores are commonly used to evaluate the performance of machine learning models, particularly for binary classification tasks~\cite{fawcett2006introduction}.
Their values range from 0 to 1, with 1 indicating perfect classification and 0.5 indicating random guessing.
Figure~\ref{fig:main_results} shows the models' AUC scores organized by misleader type and guidance level.
Across all misleaders (\autoref{fig:main_results}-A), the naive zero-shot setup yielded AUC scores above the 0.5 random guessing threshold, suggesting that the tested models can identify misleading visualizations with moderate accuracy (\textbf{RQ1}).
Of the tested models, GPT-4o generated the best overall performance, particularly in the guided zero-shot setup with an AUC score of 0.821.
We observed similar patterns for each class of misleaders.
For reasoning misleaders (\autoref{fig:main_results}-B), GPT-4o yielded the best performance in the guided few-shot setup with an AUC score of 0.835.
For design misleaders (\autoref{fig:main_results}-C), GPT-4o achieves the highest performance in the guided zero-shot setup (AUC=0.875).
Our analyses also showed that providing guidance enhances model performance in detecting misleaders across all types and levels (\textbf{RQ2}).
Moving from naive zero-shot to naive few-shot improved the AUC score for GPT-4o mini in reasoning misleaders from 0.586 to 0.811.
Explicit definitions and instructions in guided setups further boost performance, with GPT-4o's AUC score increasing from 0.806 in naive few-shot to 0.875 in guided zero-shot for design misleaders.
The most substantial gains are observed in reasoning misleaders, where the AUC score for GPT-4o rises from 0.763 in the naive few-shot to 0.835 in the guided few-shot. 

We further broke down the performance by misleader, guidance, and model (\autoref{tab:breakdown-analysis}).
The patterns were consistent with the aggregate results, indicating that tested models could identify various misleaders in the native zero-shot setting, while prompt engineering techniques improved performance.
Our findings reveal the relationship between misleader type and guidance setup.
The guided few-shot setup appeared particularly beneficial for identifying reasoning misleaders.
For instance, we found notable improvements in detecting  ``Setting an arbitrary threshold'' and ``Issues with data validity'' under the guided few-shot setup.
Conversely, the guided zero-shot setup improved the detection of design misleaders. In some cases, the AUC scores were below 0.5, suggesting the models made opposite judgments compared to the human labels.
Note that some misleaders only have a few positive and negative instances, so the AUC scores might not be reliable.

To provide insights into how the models performed the tasks, we show example responses to the visualization in the bottom-right corner of Figure~\ref{fig:teaser}, panel 1.
Due to space constraints, we only include results from GPT-4V here.
More examples can be found in the supplementary materials.
The selected visualization shows the number of lab-confirmed COVID-19 cases and the moving average in England in 2020, which decreased to less than 500 cases in May.
The accompanying tweet reads, ``\textit{This in a country of 56 million.
Lift lockdown now; the virus is just gone.}'' Lisnic et al.~\cite{lisnic2023misleading} labeled this as ``setting an arbitrary threshold,'' a reasoning misleader.
Higher-resolution versions of the figure are available in the supplementary materials.
In all setups except for guided few-shot, GPT-4V accurately described the figure and tweet text, noting that ``\textit{the decision to lift lockdown is complex and involves various factors.}''
These responses highlight GPT-4V's ability to understand visual and text inputs and its extensive knowledge of real-world events.
In the naive zero-shot setup, GPT-4V gave a certainty score of 0 for the ``setting an arbitrary threshold'' misleader (false negative), missing the issue.
GPT-4V again gave a certainty score of 0 with the naive few-shot setup, failing to detect the caption's implied threshold, stating, ``\textit{the caption suggests an opinion about lifting lockdown measures, but this does not involve setting an arbitrary threshold within the data itself.}''
In guided zero-shot, GPT-4V gave a certainty score of 100 and correctly identified that ``\textit{the caption implies a threshold at which the virus is considered `gone,' but no such threshold is indicated on the chart.}''
This demonstrates how, depending on the circumstance, certain forms of guidance may benefit GPT-4V's performance in detecting misleaders more than others.
Finally, the model gave a certainty score of 100 in the guided few-shot setup, but it mistook an example for the target and arrived at the correct conclusion by chance.
Although this mistake was uncommon, it showed that long and complex inputs could overwhelm the models, hindering performance.

\section{Discussion}
\label{sec:discussion}
We started this project with a crucial inquiry: ``Can GPT-4 models, a family of well-known LVLMs, accurately detect misleading visualizations?'' The results of our four experiments with three versions of the model (4V, o, o-mini) confirm the feasibility of using them, and possibly other comparable models, as a complementary method to combat misleading visualizations alongside other interventions, such as education.
Our findings lay the groundwork for further exploration and raise several new questions.
An immediate avenue of investigation is comparing OpenAI models with other LVLMs, such as Google's Gemini Pro~\cite{GeminiGo15:online} and LLaVA~\cite{liu2023improved} models. Additionally, our study focuses on a subset of misleaders, highlighting the need for further research to broaden the investigation and deepen our understanding across a wider array of scenarios and contexts.

Our findings reveal the relationship between misleader type and prompting strategy.
Providing richer guidance under the guided few-shot setup improved GPT-4o's performance in detecting reasoning misleaders. However, the same strategy was less effective in detecting design misleaders compared to the guided zero-shot approach, indicating that a single prompt engineering technique does not necessarily yield the best results for all types of misleaders. One possible explanation is that design misleaders heavily rely on visual patterns and structural cues, which are effectively captured through the clear and concise definitions provided in the guided zero-shot setup. The additional examples in the guided few-shot setup may introduce unnecessary complexity, potentially leading to model overfitting or causing distraction. More research is required to investigate the relationship between prompting strategies, misleader types, and model performance.

Future research must also examine and understand the models' ``reasoning strategies'' that lead to errors, as they might have adverse effects. Deeper insights can help optimize model performance and build trust in AI's capability to detect misleading visualizations and combat misinformation. Additionally, we need to explore how to best communicate the LVLMs' output to people and determine whether the systems should merely warn the users or also suggest corrected versions of the visualizations when possible. 

In our experiments, we prompted the models to examine the input for a single misleader at a time.
Since one visualization might include different misleaders simultaneously, we had to query the models multiple times to capture all of them, which is neither cost-efficient nor scalable for real-world deployment.
At the same time, our results suggest that comprehensive prompts instructing the models to detect multiple misleaders simultaneously could lead to overly lengthy inputs and outputs, potentially overwhelming the models.
Therefore, future studies should explore more effective strategies for detecting multiple misleaders.
Additionally, developing methods to efficiently balance the depth of analysis with the manageability of input complexity will be crucial.
This includes investigating hierarchical or modular prompting techniques and integrating feedback mechanisms to refine detection accuracy. 

\section{Conclusion}
In this work, we showed the potential of three GPT-4 models in identifying misleading visualizations and the efficacy of prompt engineering techniques in improving their performance.
These findings encourage further exploration into the optimal use of LVLMs for visual misinformation detection and underscore the importance of refining prompts to maximize a model's effectiveness in detecting misleaders. 

\bibliographystyle{abbrv-doi-hyperref}

\bibliography{template}

\begin{thebibliography}{10}

\bibitem{bergstrom2021calling}
C.~T. Bergstrom and J.~D. West.
\newblock {\em Calling bullshit: The art of skepticism in a data-driven world}.
\newblock Random House Trade Paperbacks, 2021. \href{https://doi.org/10.1080/1369118X.2016.1153126}
{doi: {{%
10\hspace{.1pt}\discretionary{.}{%
}{.}\hspace{.4pt}1080\discretionary{/}{%
}{/}1369118X\hspace{.1pt}\discretionary{.}{%
}{.}\hspace{.4pt}2016\hspace{.1pt}\discretionary{.}{%
}{.}\hspace{.4pt}1153126}}}


\bibitem{brahma2023leveraging}
D.~Brahma, A.~Bhattacharya, S.~N. Mahadev, A.~Asati, V.~Verma, and S.~Biswas.
\newblock Leveraging out-of-domain data for domain-specific prompt tuning in multi-modal fake news detection.
\newblock {\em arXiv preprint:2311.16496}, 2023. \href{https://doi.org/10.48550/arXiv.2311.16496}
{doi: {{%
10\hspace{.1pt}\discretionary{.}{%
}{.}\hspace{.4pt}48550\discretionary{/}{%
}{/}arXiv\hspace{.1pt}\discretionary{.}{%
}{.}\hspace{.4pt}2311\hspace{.1pt}\discretionary{.}{%
}{.}\hspace{.4pt}16496}}}


\bibitem{brown2023social}
J.~J. Brown.
\newblock Social media reflection assignment: a simple classroom intervention to help students examine scientific claims in social media.
\newblock {\em Journal of Microbiology \& Biology Education}, 22(1):55--22, 2023. \href{https://doi.org/10.1128/jmbe.00155-22}
{doi: {{%
10\hspace{.1pt}\discretionary{.}{%
}{.}\hspace{.4pt}1128\discretionary{/}{%
}{/}jmbe\hspace{.1pt}\discretionary{.}{%
}{.}\hspace{.4pt}00155\discretionary{%
}{-}{-}22}}}


\bibitem{brown2020language}
T.~Brown, B.~Mann, N.~Ryder, M.~Subbiah, J.~D. Kaplan, P.~Dhariwal, A.~Neelakantan, P.~Shyam, G.~Sastry, A.~Askell, et~al.
\newblock Language models are few-shot learners.
\newblock {\em Advances in neural information processing systems}, 33:1877--1901, 2020. \href{https://doi.org/10.48550/arXiv.2005.14165}
{doi: {{%
10\hspace{.1pt}\discretionary{.}{%
}{.}\hspace{.4pt}48550\discretionary{/}{%
}{/}arXiv\hspace{.1pt}\discretionary{.}{%
}{.}\hspace{.4pt}2005\hspace{.1pt}\discretionary{.}{%
}{.}\hspace{.4pt}14165}}}


\bibitem{cairo2019charts}
A.~Cairo.
\newblock {\em How charts lie: Getting smarter about visual information}.
\newblock WW Norton \& Company, 2019. \href{https://doi.org/10.1145/3334480.3334503}
{doi: {{%
10\hspace{.1pt}\discretionary{.}{%
}{.}\hspace{.4pt}1145\discretionary{/}{%
}{/}3334480\hspace{.1pt}\discretionary{.}{%
}{.}\hspace{.4pt}3334503}}}


\bibitem{correll2019ethical}
M.~Correll.
\newblock Ethical dimensions of visualization research.
\newblock In {\em Proceedings of the 2019 CHI conference on human factors in computing systems}, pp. 1--13, 2019. \href{https://doi.org/10.1145/3290605.3300418}
{doi: {{%
10\hspace{.1pt}\discretionary{.}{%
}{.}\hspace{.4pt}1145\discretionary{/}{%
}{/}3290605\hspace{.1pt}\discretionary{.}{%
}{.}\hspace{.4pt}3300418}}}


\bibitem{do2023llms}
X.~L. Do, M.~Hassanpour, A.~Masry, P.~Kavehzadeh, E.~Hoque, and S.~Joty.
\newblock Do {LLMs} work on charts? designing few-shot prompts for chart question answering and summarization.
\newblock {\em arXiv preprint:2312.10610}, 2023. \href{https://doi.org/10.48550/arXiv.2312.10610}
{doi: {{%
10\hspace{.1pt}\discretionary{.}{%
}{.}\hspace{.4pt}48550\discretionary{/}{%
}{/}arXiv\hspace{.1pt}\discretionary{.}{%
}{.}\hspace{.4pt}2312\hspace{.1pt}\discretionary{.}{%
}{.}\hspace{.4pt}10610}}}


\bibitem{fawcett2006introduction}
T.~Fawcett.
\newblock An introduction to roc analysis.
\newblock {\em Pattern recognition letters}, 27(8):861--874, 2006. \href{https://doi.org/10.1016/j.patrec.2005.10.010}
{doi: {{%
10\hspace{.1pt}\discretionary{.}{%
}{.}\hspace{.4pt}1016\discretionary{/}{%
}{/}j\hspace{.1pt}\discretionary{.}{%
}{.}\hspace{.4pt}patrec\hspace{.1pt}\discretionary{.}{%
}{.}\hspace{.4pt}2005\hspace{.1pt}\discretionary{.}{%
}{.}\hspace{.4pt}10\hspace{.1pt}\discretionary{.}{%
}{.}\hspace{.4pt}010}}}


\bibitem{galesic2010statistical}
M.~Galesic and R.~Garcia-Retamero.
\newblock Statistical numeracy for health: a cross-cultural comparison with probabilistic national samples.
\newblock {\em Archives of internal medicine}, 170(5):462--468, 2010. \href{https://doi.org/10.1001/archinternmed.2010.11}
{doi: {{%
10\hspace{.1pt}\discretionary{.}{%
}{.}\hspace{.4pt}1001\discretionary{/}{%
}{/}archinternmed\hspace{.1pt}\discretionary{.}{%
}{.}\hspace{.4pt}2010\hspace{.1pt}\discretionary{.}{%
}{.}\hspace{.4pt}11}}}


\bibitem{GeminiGo15:online}
{Google DeepMind}.
\newblock Gemini.
\newblock \url{https://deepmind.google/technologies/gemini/\#introduction}, 2024.
\newblock (Accessed on 04/29/2024).

\bibitem{huang2023lvlms}
K.-H. Huang, M.~Zhou, H.~P. Chan, Y.~R. Fung, Z.~Wang, L.~Zhang, S.-F. Chang, and H.~Ji.
\newblock Do {LVLMs} understand charts? {A}nalyzing and correcting factual errors in chart captioning.
\newblock {\em arXiv preprint:2312.10160}, 2023. \href{https://doi.org/10.48550/arXiv.2312.10160}
{doi: {{%
10\hspace{.1pt}\discretionary{.}{%
}{.}\hspace{.4pt}48550\discretionary{/}{%
}{/}arXiv\hspace{.1pt}\discretionary{.}{%
}{.}\hspace{.4pt}2312\hspace{.1pt}\discretionary{.}{%
}{.}\hspace{.4pt}10160}}}


\bibitem{jones2004models}
G.~A. Jones, C.~W. Langrall, E.~S. Mooney, and C.~A. Thornton.
\newblock Models of development in statistical reasoning.
\newblock {\em The challenge of developing statistical literacy, reasoning and thinking}, pp. 97--117, 2004. \href{https://doi.org/10.1007/978-1-4020-2278-6_5}
{doi: {{%
10\hspace{.1pt}\discretionary{.}{%
}{.}\hspace{.4pt}1007\discretionary{/}{%
}{/}978\discretionary{%
}{-}{-}1\discretionary{%
}{-}{-}4020\discretionary{%
}{-}{-}2278\discretionary{%
}{-}{-}6\_5}}}


\bibitem{jung2017chartsense}
D.~Jung, W.~Kim, H.~Song, J.-i. Hwang, B.~Lee, B.~Kim, and J.~Seo.
\newblock Chartsense: Interactive data extraction from chart images.
\newblock In {\em Proceedings of the 2017 chi conference on human factors in computing systems}, pp. 6706--6717, 2017. \href{https://doi.org/10.1145/3025453.3025957}
{doi: {{%
10\hspace{.1pt}\discretionary{.}{%
}{.}\hspace{.4pt}1145\discretionary{/}{%
}{/}3025453\hspace{.1pt}\discretionary{.}{%
}{.}\hspace{.4pt}3025957}}}


\bibitem{kennedy2016work}
H.~Kennedy, R.~L. Hill, G.~Aiello, and W.~Allen.
\newblock The work that visualisation conventions do.
\newblock {\em Information, Communication \& Society}, 19(6):715--735, 2016. \href{https://doi.org/10.1080/1369118X.2016.1153126}
{doi: {{%
10\hspace{.1pt}\discretionary{.}{%
}{.}\hspace{.4pt}1080\discretionary{/}{%
}{/}1369118X\hspace{.1pt}\discretionary{.}{%
}{.}\hspace{.4pt}2016\hspace{.1pt}\discretionary{.}{%
}{.}\hspace{.4pt}1153126}}}


\bibitem{kim2023good}
N.~W. Kim, G.~Myers, and B.~Bach.
\newblock How good is {ChatGPT} in giving advice on your visualization design?
\newblock {\em arXiv preprint:2310.09617}, 2023. \href{https://doi.org/10.48550/arXiv.2310.09617}
{doi: {{%
10\hspace{.1pt}\discretionary{.}{%
}{.}\hspace{.4pt}48550\discretionary{/}{%
}{/}arXiv\hspace{.1pt}\discretionary{.}{%
}{.}\hspace{.4pt}2310\hspace{.1pt}\discretionary{.}{%
}{.}\hspace{.4pt}09617}}}


\bibitem{kruger1999unskilled}
J.~Kruger and D.~Dunning.
\newblock Unskilled and unaware of it: how difficulties in recognizing one's own incompetence lead to inflated self-assessments.
\newblock {\em Journal of personality and social psychology}, 77(6):1121, 1999. \href{https://doi.org/10.1037/0022-3514.77.6.1121}
{doi: {{%
10\hspace{.1pt}\discretionary{.}{%
}{.}\hspace{.4pt}1037\discretionary{/}{%
}{/}0022\discretionary{%
}{-}{-}3514\hspace{.1pt}\discretionary{.}{%
}{.}\hspace{.4pt}77\hspace{.1pt}\discretionary{.}{%
}{.}\hspace{.4pt}6\hspace{.1pt}\discretionary{.}{%
}{.}\hspace{.4pt}1121}}}


\bibitem{langraw2023study}
K.~S. Langraw and A.~Zaman.
\newblock A study on evaluating the impact of social media's fake news on the attitudes and beliefs of a society.
\newblock {\em International Journal of Social Science \& Entrepreneurship}, 3(4):254--270, 2023. \href{https://doi.org/10.58661/ijsse.v3i4.254}
{doi: {{%
10\hspace{.1pt}\discretionary{.}{%
}{.}\hspace{.4pt}58661\discretionary{/}{%
}{/}ijsse\hspace{.1pt}\discretionary{.}{%
}{.}\hspace{.4pt}v3i4\hspace{.1pt}\discretionary{.}{%
}{.}\hspace{.4pt}254}}}


\bibitem{lin2021fooled}
C.~Lin and M.~A. Thornton.
\newblock Fooled by beautiful data: Visualization aesthetics bias trust in science, news, and social media.
\newblock {\em PsyArXiv}, 2021. \href{https://doi.org/10.31234/osf.io/h8tc2}
{doi: {{%
10\hspace{.1pt}\discretionary{.}{%
}{.}\hspace{.4pt}31234\discretionary{/}{%
}{/}osf\hspace{.1pt}\discretionary{.}{%
}{.}\hspace{.4pt}io\discretionary{/}{%
}{/}h8tc2}}}


\bibitem{lin2023beneath}
H.~Lin, Z.~Luo, J.~Ma, and L.~Chen.
\newblock Beneath the surface: Unveiling harmful memes with multimodal reasoning distilled from large language models.
\newblock {\em arXiv preprint:2312.05434}, 2023. \href{https://doi.org/10.48550/arXiv.2312.05434}
{doi: {{%
10\hspace{.1pt}\discretionary{.}{%
}{.}\hspace{.4pt}48550\discretionary{/}{%
}{/}arXiv\hspace{.1pt}\discretionary{.}{%
}{.}\hspace{.4pt}2312\hspace{.1pt}\discretionary{.}{%
}{.}\hspace{.4pt}05434}}}


\bibitem{lisnic2023yeah}
M.~Lisnic, A.~Lex, and M.~Kogan.
\newblock ``{Yeah}, this graph doesn't show that'': Analysis of online engagement with misleading data visualizations.
\newblock {\em OSF Preprints}, 2023. \href{https://doi.org/10.31219/osf.io/s5v6x}
{doi: {{%
10\hspace{.1pt}\discretionary{.}{%
}{.}\hspace{.4pt}31219\discretionary{/}{%
}{/}osf\hspace{.1pt}\discretionary{.}{%
}{.}\hspace{.4pt}io\discretionary{/}{%
}{/}s5v6x}}}


\bibitem{lisnic2023misleading}
M.~Lisnic, C.~Polychronis, A.~Lex, and M.~Kogan.
\newblock Misleading beyond visual tricks: How people actually lie with charts.
\newblock In {\em Proceedings of the 2023 CHI Conference on Human Factors in Computing Systems}, pp. 1--21, 2023. \href{https://doi.org/10.1145/3544548.3581087}
{doi: {{%
10\hspace{.1pt}\discretionary{.}{%
}{.}\hspace{.4pt}1145\discretionary{/}{%
}{/}3544548\hspace{.1pt}\discretionary{.}{%
}{.}\hspace{.4pt}3581087}}}


\bibitem{liu2023improved}
H.~Liu, C.~Li, Y.~Li, and Y.~J. Lee.
\newblock Improved baselines with visual instruction tuning.
\newblock {\em arXiv preprint:2310.03744}, 2023. \href{https://doi.org/10.48550/arXiv.2310.03744}
{doi: {{%
10\hspace{.1pt}\discretionary{.}{%
}{.}\hspace{.4pt}48550\discretionary{/}{%
}{/}arXiv\hspace{.1pt}\discretionary{.}{%
}{.}\hspace{.4pt}2310\hspace{.1pt}\discretionary{.}{%
}{.}\hspace{.4pt}03744}}}


\bibitem{lo2022misinformed}
L.~Y.-H. Lo, A.~Gupta, K.~Shigyo, A.~Wu, E.~Bertini, and H.~Qu.
\newblock Misinformed by visualization: What do we learn from misinformative visualizations?
\newblock In {\em Computer Graphics Forum}, vol.~41, pp. 515--525. Wiley Online Library, 2022. \href{https://doi.org/10.1111/cgf.14500}
{doi: {{%
10\hspace{.1pt}\discretionary{.}{%
}{.}\hspace{.4pt}1111\discretionary{/}{%
}{/}cgf\hspace{.1pt}\discretionary{.}{%
}{.}\hspace{.4pt}14500}}}


\bibitem{loomba2021measuring}
S.~Loomba, A.~de~Figueiredo, S.~J. Piatek, K.~de~Graaf, and H.~J. Larson.
\newblock Measuring the impact of covid-19 vaccine misinformation on vaccination intent in the uk and usa.
\newblock {\em Nature human behaviour}, 5(3):337--348, 2021. \href{https://doi.org/10.1038/s41562-021-01056-1}
{doi: {{%
10\hspace{.1pt}\discretionary{.}{%
}{.}\hspace{.4pt}1038\discretionary{/}{%
}{/}s41562\discretionary{%
}{-}{-}021\discretionary{%
}{-}{-}01056\discretionary{%
}{-}{-}1}}}


\bibitem{GPT4Visi34:online}
{OpenAI}.
\newblock {GPT-4V(ision)} system card.
\newblock \url{https://openai.com/research/gpt-4v-system-card}, Sep 2023.
\newblock (Accessed on 02/02/2024).

\bibitem{GPT4omini:online}
{OpenAI}.
\newblock {GPT-4o mini: advancing cost-efficient intelligence}.
\newblock \url{https://openai.com/index/gpt-4o-mini-advancing-cost-efficient-intelligence}, July 2024.
\newblock (Accessed on 08/03/2024).

\bibitem{GPT4o:online}
{OpenAI}.
\newblock {Hello GPT-4o}.
\newblock \url{https://openai.com/index/hello-gpt-4o}, May 2024.
\newblock (Accessed on 08/03/2024).

\bibitem{Prompten19:online}
{OpenAI API}.
\newblock Prompt engineering.
\newblock \url{https://platform.openai.com/docs/guides/prompt-engineering}.
\newblock (Accessed on 04/30/2024).

\bibitem{pennycook2020fighting}
G.~Pennycook, J.~McPhetres, Y.~Zhang, J.~G. Lu, and D.~G. Rand.
\newblock Fighting covid-19 misinformation on social media: Experimental evidence for a scalable accuracy-nudge intervention.
\newblock {\em Psychological science}, 31(7):770--780, 2020. \href{https://doi.org/10.1177/0956797620939054}
{doi: {{%
10\hspace{.1pt}\discretionary{.}{%
}{.}\hspace{.4pt}1177\discretionary{/}{%
}{/}0956797620939054}}}


\bibitem{rivera2022social}
O.~Rivera-Romero, E.~Gabarron, T.~Miron-Shatz, C.~Petersen, and K.~Denecke.
\newblock Social media, digital health literacy, and digital ethics in the light of health equity.
\newblock {\em Yearbook of Medical Informatics}, 31(01):082--087, 2022. \href{https://doi.org/10.1055/s-0042-1742395}
{doi: {{%
10\hspace{.1pt}\discretionary{.}{%
}{.}\hspace{.4pt}1055\discretionary{/}{%
}{/}s\discretionary{%
}{-}{-}0042\discretionary{%
}{-}{-}1742395}}}


\bibitem{shao2018spread}
C.~Shao, G.~L. Ciampaglia, O.~Varol, K.-C. Yang, A.~Flammini, and F.~Menczer.
\newblock The spread of low-credibility content by social bots.
\newblock {\em Nature communications}, 9(1):1--9, 2018. \href{https://doi.org/10.1038/s41467-018-06930-7}
{doi: {{%
10\hspace{.1pt}\discretionary{.}{%
}{.}\hspace{.4pt}1038\discretionary{/}{%
}{/}s41467\discretionary{%
}{-}{-}018\discretionary{%
}{-}{-}06930\discretionary{%
}{-}{-}7}}}


\bibitem{simons1999gorillas}
D.~J. Simons and C.~F. Chabris.
\newblock Gorillas in our midst: Sustained inattentional blindness for dynamic events.
\newblock {\em Perception}, 28(9):1059--1074, 1999. \href{https://doi.org/10.1068/p281059}
{doi: {{%
10\hspace{.1pt}\discretionary{.}{%
}{.}\hspace{.4pt}1068\discretionary{/}{%
}{/}p281059}}}


\bibitem{sweller1988cognitive}
J.~Sweller.
\newblock Cognitive load during problem solving: Effects on learning.
\newblock {\em Cognitive science}, 12(2):257--285, 1988. \href{https://doi.org/10.1207/s15516709cog1202_4}
{doi: {{%
10\hspace{.1pt}\discretionary{.}{%
}{.}\hspace{.4pt}1207\discretionary{/}{%
}{/}s15516709cog1202\_4}}}


\bibitem{van2017inoculating}
S.~Van~der Linden, A.~Leiserowitz, S.~Rosenthal, and E.~Maibach.
\newblock Inoculating the public against misinformation about climate change.
\newblock {\em Global challenges}, 1(2):1600008, 2017. \href{https://doi.org/10.1002/gch2.201600008}
{doi: {{%
10\hspace{.1pt}\discretionary{.}{%
}{.}\hspace{.4pt}1002\discretionary{/}{%
}{/}gch2\hspace{.1pt}\discretionary{.}{%
}{.}\hspace{.4pt}201600008}}}


\bibitem{vazquez2024llms}
P.-P. V{\'a}zquez.
\newblock Are {LLMs} ready for visualization?
\newblock {\em arXiv preprint:2403.06158}, 2024. \href{https://doi.org/10.48550/arXiv.2403.06158}
{doi: {{%
10\hspace{.1pt}\discretionary{.}{%
}{.}\hspace{.4pt}48550\discretionary{/}{%
}{/}arXiv\hspace{.1pt}\discretionary{.}{%
}{.}\hspace{.4pt}2403\hspace{.1pt}\discretionary{.}{%
}{.}\hspace{.4pt}06158}}}


\bibitem{vergho2024comparing}
T.~Vergho, J.-F. Godbout, R.~Rabbany, and K.~Pelrine.
\newblock Comparing {GPT}-4 and open-source language models in misinformation mitigation.
\newblock {\em arXiv preprint:2401.06920}, 2024. \href{https://doi.org/10.48550/arXiv.2401.06920}
{doi: {{%
10\hspace{.1pt}\discretionary{.}{%
}{.}\hspace{.4pt}48550\discretionary{/}{%
}{/}arXiv\hspace{.1pt}\discretionary{.}{%
}{.}\hspace{.4pt}2401\hspace{.1pt}\discretionary{.}{%
}{.}\hspace{.4pt}06920}}}


\bibitem{vosoughi2018spread}
S.~Vosoughi, D.~Roy, and S.~Aral.
\newblock The spread of true and false news online.
\newblock {\em science}, 359(6380):1146--1151, 2018. \href{https://doi.org/10.1126/science.aap9559}
{doi: {{%
10\hspace{.1pt}\discretionary{.}{%
}{.}\hspace{.4pt}1126\discretionary{/}{%
}{/}science\hspace{.1pt}\discretionary{.}{%
}{.}\hspace{.4pt}aap9559}}}


\bibitem{vrabec2023popularisation}
N.~Vrabec and L.~Pie{\v{s}}.
\newblock Popularisation of science and science journalism on social media in slovakia.
\newblock {\em Media Literacy and Academic Research}, 6(1):206--226, 2023. \href{https://doi.org/10.17846/MLAR.2023.6.1.206-226}
{doi: {{%
10\hspace{.1pt}\discretionary{.}{%
}{.}\hspace{.4pt}17846\discretionary{/}{%
}{/}MLAR\hspace{.1pt}\discretionary{.}{%
}{.}\hspace{.4pt}2023\hspace{.1pt}\discretionary{.}{%
}{.}\hspace{.4pt}6\hspace{.1pt}\discretionary{.}{%
}{.}\hspace{.4pt}1\hspace{.1pt}\discretionary{.}{%
}{.}\hspace{.4pt}206\discretionary{%
}{-}{-}226}}}


\bibitem{yang2024empowering}
Y.~Yang, X.~Zhang, J.~Xu, and W.~Han.
\newblock Empowering vision-language models for reasoning ability through large language models.
\newblock In {\em ICASSP 2024-2024 IEEE International Conference on Acoustics, Speech and Signal Processing (ICASSP)}, pp. 10056--10060. IEEE, 2024. \href{https://doi.org/10.1109/ICASSP.2024.9413054}
{doi: {{%
10\hspace{.1pt}\discretionary{.}{%
}{.}\hspace{.4pt}1109\discretionary{/}{%
}{/}ICASSP\hspace{.1pt}\discretionary{.}{%
}{.}\hspace{.4pt}2024\hspace{.1pt}\discretionary{.}{%
}{.}\hspace{.4pt}9413054}}}


\bibitem{zakir2023infodemics}
I.~Zakir~Hussain, J.~Kaur, M.~Lotto, Z.~Butt, and P.~Morita.
\newblock Infodemics surveillance system to detect and analyze health misinformation using big data and {AI}.
\newblock {\em European Journal of Public Health}, 33(Supplement\_2):ckad160--163, 2023. \href{https://doi.org/10.1093/eurpub/ckad160}
{doi: {{%
10\hspace{.1pt}\discretionary{.}{%
}{.}\hspace{.4pt}1093\discretionary{/}{%
}{/}eurpub\discretionary{/}{%
}{/}ckad160}}}


\bibitem{zhang2023detecting}
Y.~Zhang, L.~Trinh, D.~Cao, Z.~Cui, and Y.~Liu.
\newblock Detecting out-of-context multimodal misinformation with interpretable neural-symbolic model.
\newblock {\em arXiv preprint:2304.07633}, 2023. \href{https://doi.org/10.48550/arXiv.2304.07633}
{doi: {{%
10\hspace{.1pt}\discretionary{.}{%
}{.}\hspace{.4pt}48550\discretionary{/}{%
}{/}arXiv\hspace{.1pt}\discretionary{.}{%
}{.}\hspace{.4pt}2304\hspace{.1pt}\discretionary{.}{%
}{.}\hspace{.4pt}07633}}}


\end{thebibliography}

\end{document}